\documentclass[conference]{IEEEtran}
\IEEEoverridecommandlockouts
\usepackage{cite}
\usepackage{amsmath,amssymb,amsfonts}
\usepackage{algorithmic}
\usepackage{graphicx}
\usepackage{textcomp}
\usepackage{xcolor}
\def\BibTeX{{\rm B\kern-.05em{\sc i\kern-.025em b}\kern-.08em
    T\kern-.1667em\lower.7ex\hbox{E}\kern-.125emX}}
\begin{document}

\title{Cloud-Based Face and Speech Recognition for Access Control Applications
\thanks{This work has been carried out by Nathalie Tkauc and Thao Tran in the context of their Bachelor Thesis at Halmstad University.}
}

\author{\IEEEauthorblockN{Nathalie Tkauc, Thao Tran, Kevin Hernandez-Diaz, Fernando Alonso-Fernandez}
\IEEEauthorblockA{\textit{School of Information Technology (ITE)} \\
\textit{Halmstad University}, Sweden \\
kevin.hernandez-diaz@hh.se, feralo@hh.se}
}

\maketitle

\begin{abstract}
This paper describes the implementation of a system to recognize
employees and visitors wanting to gain access to a physical office
through face images and speech-to-text recognition.
The system helps employees to unlock the entrance door via face recognition
without the need of tag-keys or cards.
To prevent spoofing attacks and increase security,
a randomly generated code is sent to the employee,
who then has to type it into the screen.
On the other hand,
visitors and delivery persons are provided with a
speech-to-text service where they utter the name
of the employee that they want to meet, and the system
then sends a notification to the right employee automatically.
The hardware of the system is constituted by two Raspberry Pi, a 7-inch LCD-touch
display, a camera, and a sound card with a microphone and speaker.
To carry out face recognition and speech-to-text conversion,
the cloud-based platforms Amazon Web Services
and the Google Speech-to-Text API service are used respectively.
The two-step face authentication mechanism for employees
provides an increased level of security and protection against spoofing attacks
without the need of carrying key-tags or access cards,
while disturbances by visitors or couriers are minimized
by notifying their arrival to the right employee,
without disturbing other co-workers by means of ring-bells.
%
%
%
\end{abstract}


\section{Introduction}

This work is a collaboration with the company JayWay in Halmstad.
In order to enter the office today, employees need a tag-key.
Those who do not have a tag-key (for example visitors or couriers),
or if employees forget their tag,
they must ring the doorbell.
Then, someone has to open the door manually,
usually those at the desks located closer to the entrance.
This does not only consume time and focus of the workday for
these staff members but also disturbs others with desks
located nearby.
Accordingly, the goal of this work is to develop a system that uses face recognition to control the lock system of the entrance door for employees,
and speech-to-text conversion to notify the appropriate staff member that a visitor or delivery is waiting outside.
The project will help the company and its staff in two ways.
First, it will simplify the process of entering the office for employees.
Second, it will also reduce the disturbance when a visitor wants to enter the office
or when an employee forgets their tag-key.

\begin{figure}[htb]
\centering
        \includegraphics[width=0.48\textwidth]{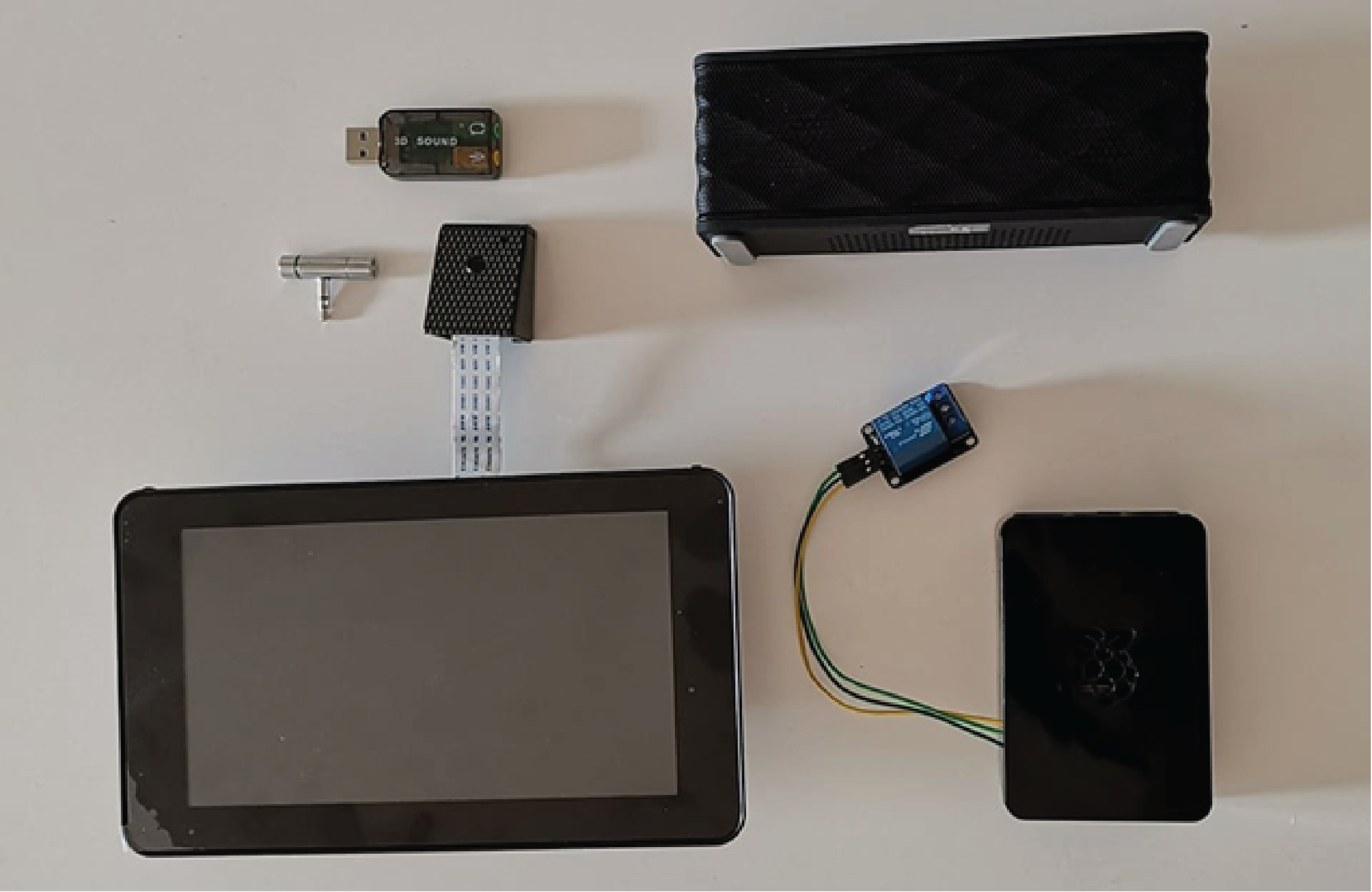}
\caption{Hardware parts used in the project.}
\label{fig:hardware}
\end{figure}

The work of this paper will be limited to the fundamentals
of the access control system.
In terms of hardware, the fundamentals
are a small computer, display, camera, mic and speaker (Figure~\ref{fig:hardware}).
%
The requirements set for this project are:
\begin{itemize}
  \item To control the access to the office with the help of a small computer and a camera.
  \item To identify employees by means of a picture that will be compared against images in the personnel database.
  \item To unlock the entrance door when an employee is granted access.
  \item To use speech-to-text to handle visitors and guests by identifying the employee that should handle the visit.
  \item To notify the employee that a visitor is waiting at the door.
\end{itemize}

\begin{figure}[htb]
\centering
        \includegraphics[width=0.35\textwidth]{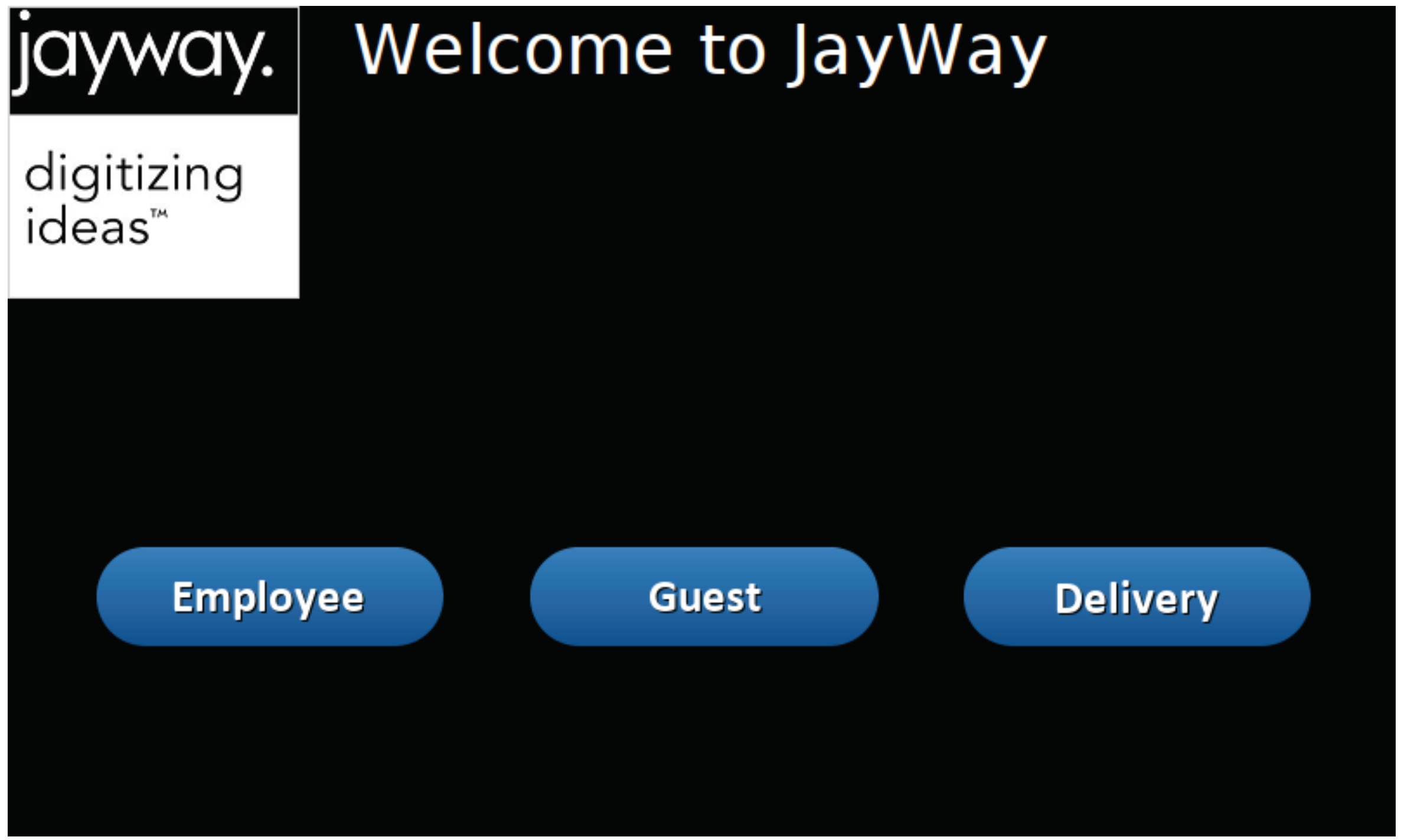}
\caption{Main menu on the GUI.}
\label{fig:gui_main_menu}
\end{figure}

A user interface has been
developed as well to handle user interaction via the display (Figure~\ref{fig:gui_main_menu}).
The display will show a graphical user interface (GUI) for the
users to interact with and present different outputs.

\subsection{Related Works}

Today there are different technologies for facial recognition,
including 2D, 2D-3D and 3D facial recognition \cite{ABATE20071885}.
The accuracy will depend on several factors including
light,
pose,
occlusion,
or expression changes \cite{SINGH2018536}.
Face recognition technologies are in extensive use today
due to social networks and the massive availability of cameras
in many personal devices.
In such contexts, cloud computing moves the computing power and data storage burden to the cloud, bringing the possibility of biometric applications to a range of devices that otherwise would not have sufficient capacity to perform the operations locally \cite{6822213}.
Solutions have been proposed for example for face recognition \cite{6844616,6920723},
iris \cite{Kesava14},
voice \cite{6138538},
or keystroke \cite{Xi}.
Cloud speech-to-text solutions are also in increasing use by millions of person in applications such as Siri, Google, Amazon Transcribe, Alexa etc.
Speech-to-text solutions provide the capability to translate verbal language to text, enabling a more natural human-machine interaction, given that voice is the most common method for people to communicate \cite{Das}.
%
%

Notifying an employee can be done in several ways, e.g. via SMS, E-mail or Slack. Slack is a tool that can be used for communication within a group, a whole office, etc.\footnote{https://slack.com/} Different channels can be created as well for different projects. The company involved in this work uses Slack to communicate within their offices. A bot can be created in the slack workspace that can be used to send messages to inform the employees regarding if there is a guest outside waiting for them, if there is a delivery at the door, etc. This bot will be referred to as Slack-bot.
Both Amazon and Google offers services to send E-mail with respective API, but only Amazon offers the API to send an SMS. However these services are pay-per-use.

Biometric systems can be attacked via spoofing attacks \cite{[Jain11a]}.
These are different from regular IT attacks, in which system channels or modules
are hijacked.
In spoofing attacks, a fake biometric characteristic is presented to the sensor
with the purpose to fool the system and pretend to be somebody else.
Besides the apparent vulnerability that this may entail,
such type of attacks does not require any knowledge or access to the inner of the system,
and they can be carried out without any engineering skills.
In the case of face systems, for example, a picture, a video, or a 3D mask of a genuine used can be held in front of the camera \cite{6990726}.
Unfortunately, today it is straightforward to get a picture of a person directly or from the internet, even inadvertently.
For this reason, it is of utmost importance to implement measures to counteract
spoofing attacks.

\begin{figure}[htb]
\centering
        \includegraphics[width=0.4\textwidth]{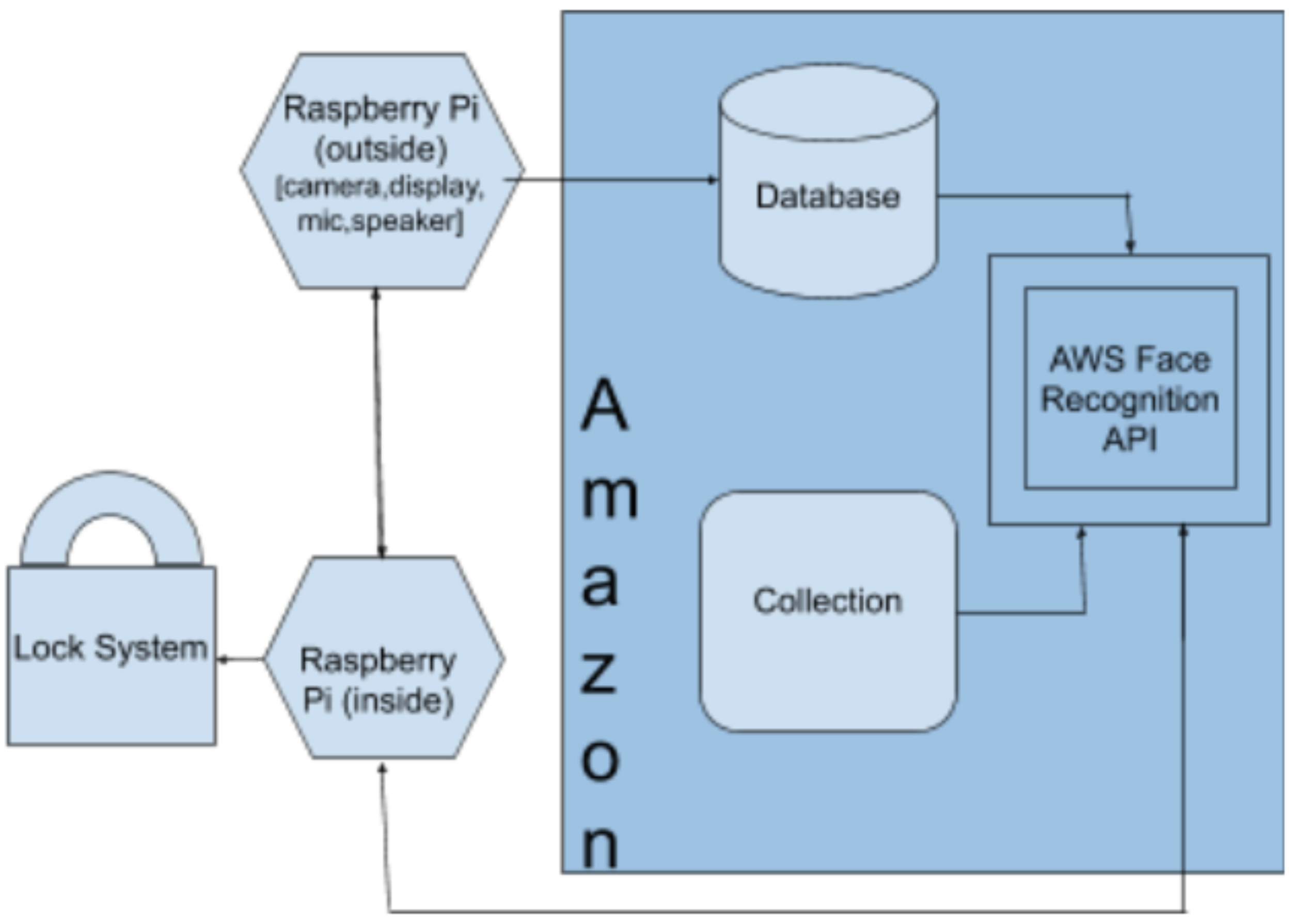}
\caption{Overview of the whole system.}
\label{fig:flow_system}
\end{figure}

\section{System Design}

The project is divided into two phases, the experimentation
phase and the implementation phase. The experimentation
phase aims to find the most suitable setup for this project.
It lays the foundation of how the system will be implemented.
The implementation phase seeks to describe how the software
and hardware are integrated to form the intended system.

\begin{figure*}[htb]
\centering
        \includegraphics[width=0.6\textwidth]{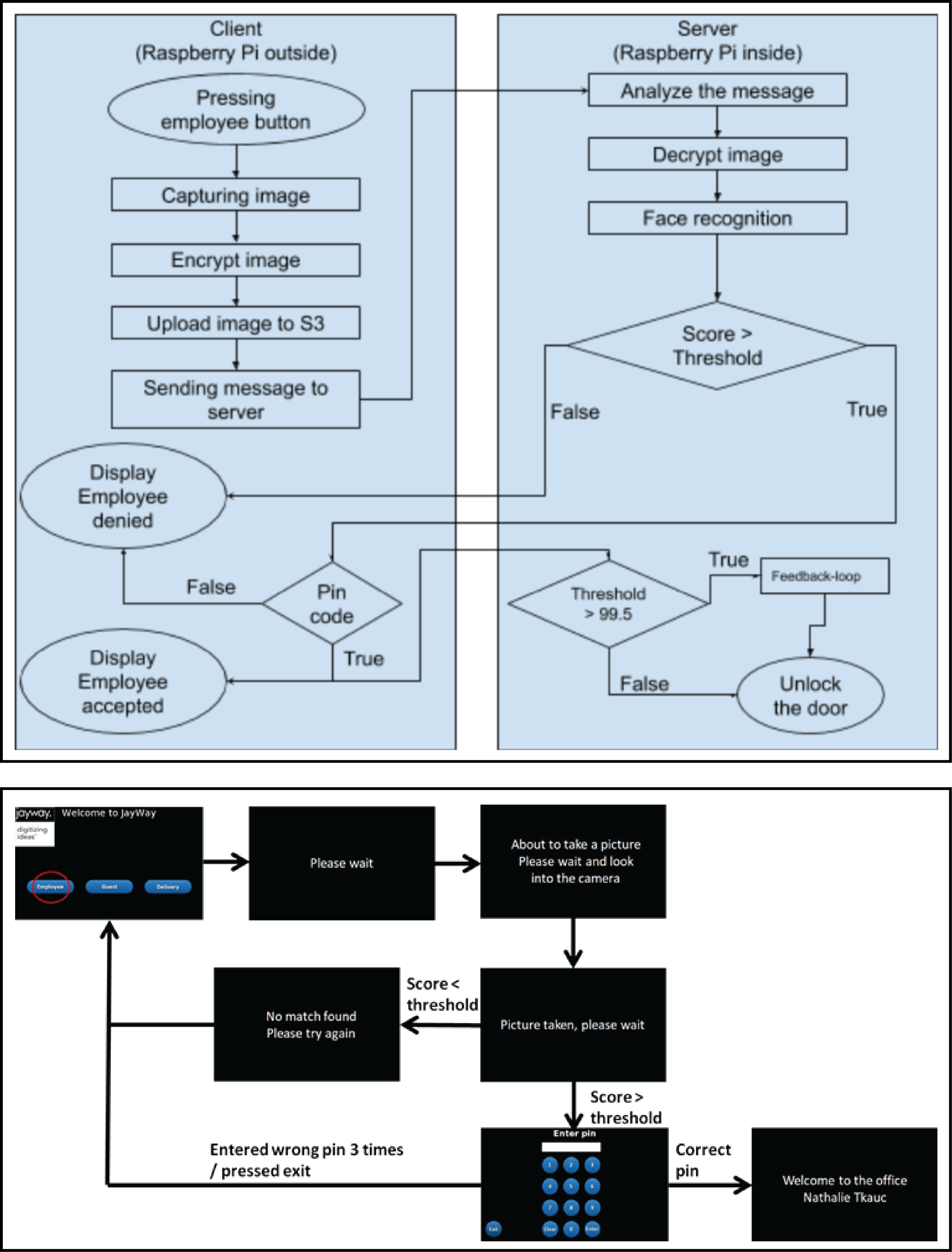}
\caption{Flowcharts of employee function (top) and GUI (bottom).}
\label{fig:flow_employee}
\end{figure*}

\subsection{Software Components}

Python v3 was chosen as the programming language. We also considered Java, but Python offered more APIs and libraries. 
Overall, Python offers a simplicity when it comes to libraries and APIs for calculations and recognitions, plus it is a well-known programming language used for image and data analysis \cite{8665702}.

Currently, the company has its information on each employee on Kperson, which is their custom-built website where they keep all the data within the company. Kperson is deployed on Amazon. The information has the form of a json string, and it contains all the necessary strings such as name, last name, id, company, link to images etc. This information can be used for face recognition, which saves the effort of creating a new database. 

For face recognition, we employed Amazon Rekognition \cite{AmazonRek} and Amazon Simple Storage Service (Amazon S3), both from Amazon Web Services (AWS).
%
Amazon Rekognition is a cloud-based software providing computer vision capabilities, while Amazon S3 provides object storage.
The reason for choosing AWS was that it provided the most important services that were needed. Another reason is also that the company had its data on Amazon.
We also considered the use of Google Cloud Platform (GCP), a
platform which provided similar services to what AWS did.
The reasons why GCP was not chosen were because it would have demanded the transfer of employee data (already at Amazon) to GCP.
Moreover, GCP did not provide a face recognition service, only face detection. 
All the data for each employee was fetched from Kperson, such like their full name and images.

Speech-to-text is used for the guest function, where the guest has to speak the name of the employee they want to meet with. Both Amazon Transcribe (AT) \cite{AmazonSpeechText} and Google's Cloud Speech-to-Text (GCST) \cite{GoogleSpeechText} services were considered. 
We have chosen Google's service, since in our tests, Amazon's engine took 30 to 60 seconds to do a translation.
This may be because AT first record a file, send it to the cloud and then translate it into a string, while GCST takes the input directly and translate it into a string much faster.
Another reason to choose Google's service is that Amazon does not support Swedish language (the working language of the company involved).

The chosen method for notifying a employee in the guest handling is Slack. Amazon offered services for both E-mail and SMS, but both costs depending on the amount of messages sent, while Slack is free. Therefore, to reduce the costs of the project and because Slack is already used as a primary communication tool within the company, Slack was chosen for the notifications. To be able to send notifications, a slack-bot was created.

\begin{figure*}[t!]
\centering
        \includegraphics[width=0.6\textwidth]{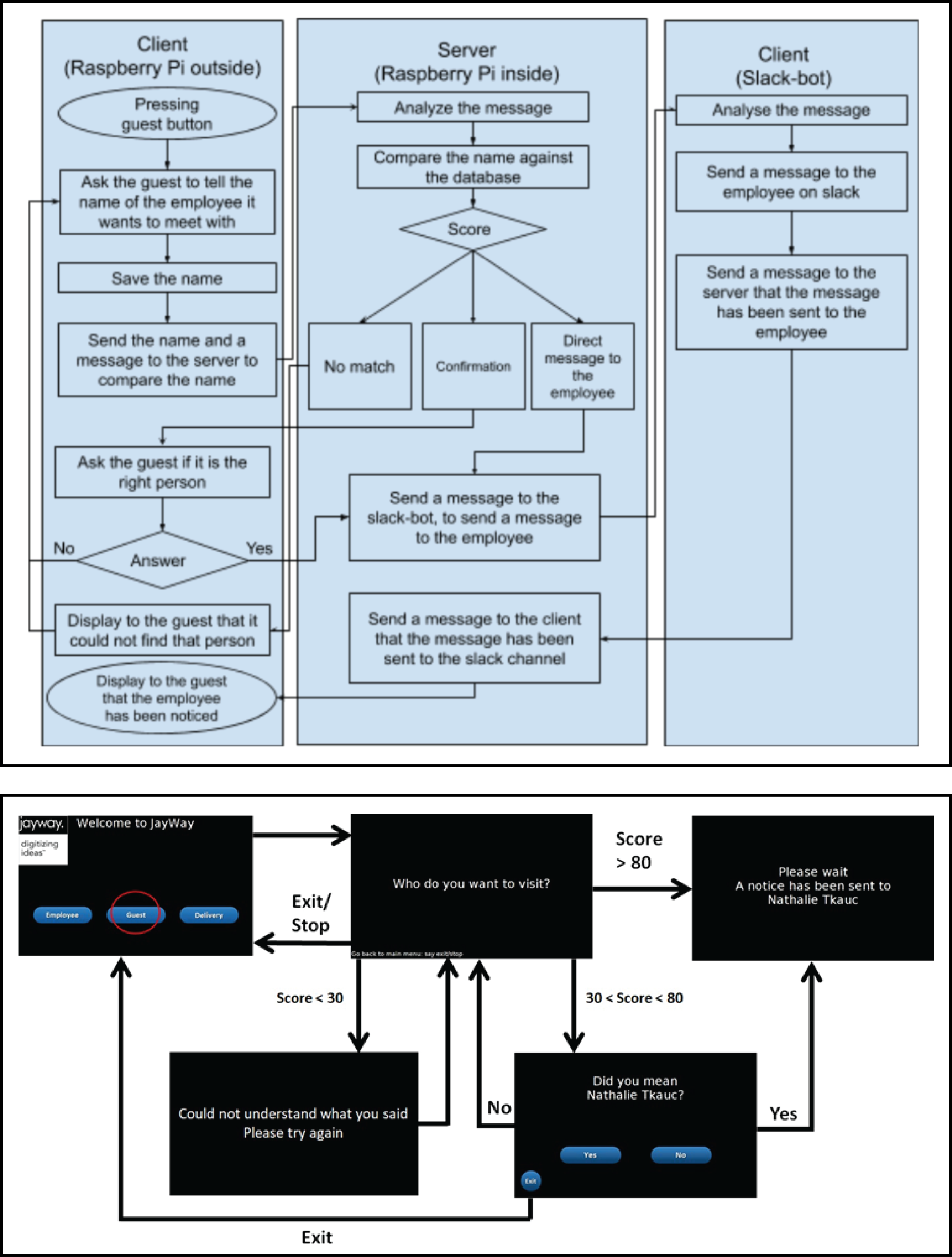}
\caption{Flowcharts of guest function (top) and GUI (bottom).}
\label{fig:flow_guest}
\end{figure*}

\subsection{Hardware Components}

The hardware-tools needed for the project are the following: two Raspberry Pi, a camera for the face recognition, a microphone for the speech-to-text services, a speaker, a touch screen to show the GUI, access to a database with the face of the company’s employees, and finally the lock system, which should be connected to a Raspberry Pi and therefore be able to be controlled by code.

There were two hardware candidates suitable for the project, either a Raspberry Pi or an Arduino. One of the most primary programming languages hosted on the Raspberry Pi is Python, while Arduino cannot run a code written in Python, but it is possible for it to communicate with a device using Python. To be able to use the cloud services, a Raspberry Pi can easily be connected to internet while a Arduino needs a external hardware. Overall the Raspberry Pi and Arduino are very similar but the Raspberry Pi are more powerful than the Arduino and is better on handling multiple tasks, making it more suitable for this project. Therefore Raspberry Pi (Model: Raspberry Pi 3 Model B+) was selected. The Raspberry Pi also has other devices that are specially adapted for it, which was used for the project and stated below.

Two devices of Raspberry Pi was used just because of security reasons (Figure~\ref{fig:flow_system}). One will be placed outside the office with a touch screen, camera, microphone and a speaker. These components only handle the capturing of the image and sending the image to the cloud storage. The second Raspberry Pi will be placed inside the office and handle the critical tasks of face recognition and controlling the lock system (this is the reason why this Raspberry Pi is placed inside the office). Both of the Raspberry Pi’s are running on the operating system called Raspbian. 
The Raspberry Pi-to-Raspberry Pi communication was done using transmission control protocol (TCP). 
The Raspberry Pi device that was placed indoors was the server and it had two clients connected to it. One client was the other Raspberry Pi, that will be mounted outside the entrance door and the other one was the slack-bot.

\begin{figure*}[t!]
\centering
        \includegraphics[width=0.6\textwidth]{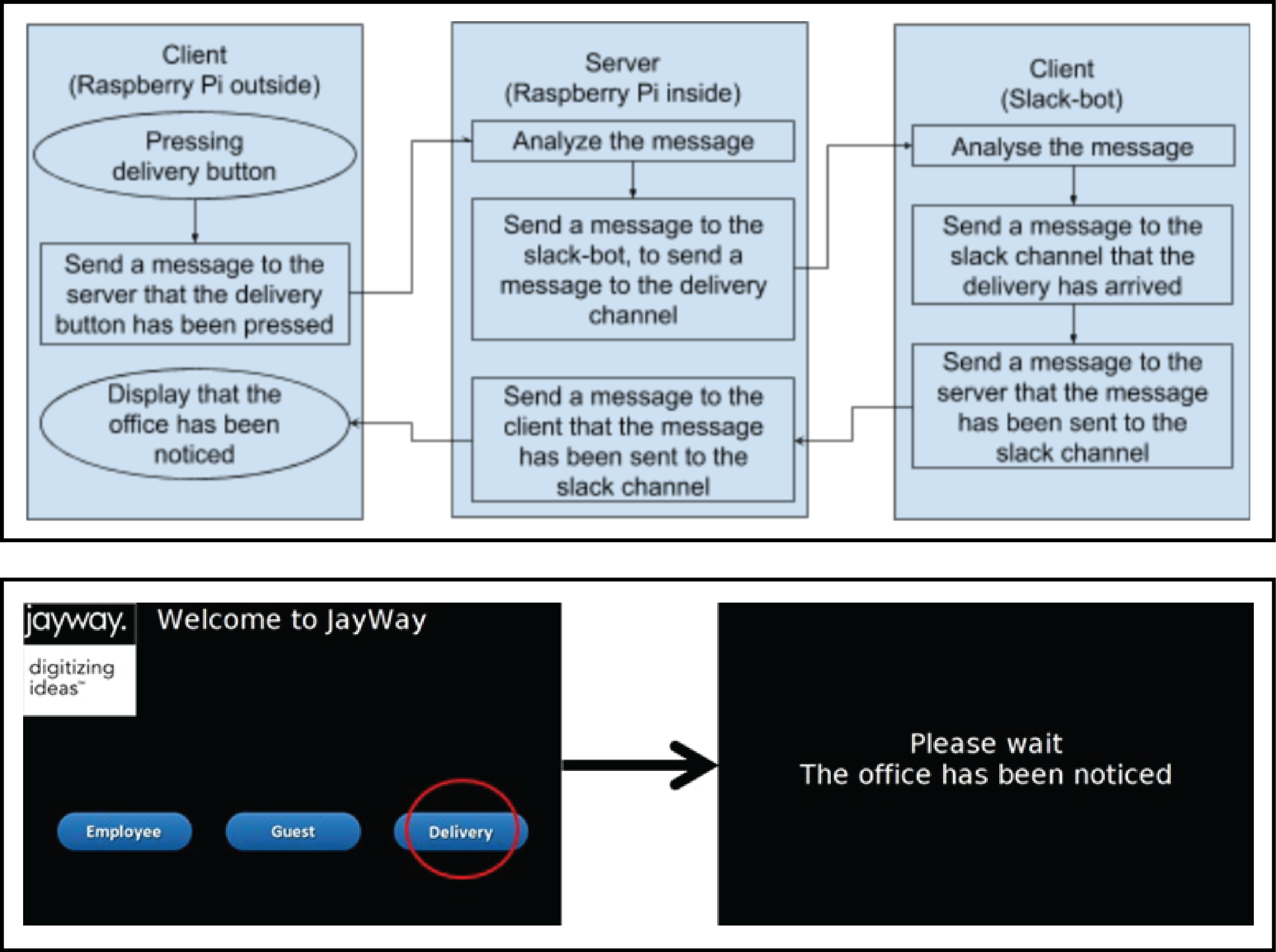}
\caption{Flowcharts of delivery function (top) and GUI (bottom).}
\label{fig:flow_delivery}
\end{figure*}

A 7 inch LCD-touch display (Model: RASPBERRYPI-DISPLAY) was attached to the device to show the GUI. It is a multi-touch capacitive touch screen and supports up to 10 fingers at the same time. The resolution on the screen is 800$\times$480 and has a update frequency on 60hz. The display needed a Raspberry Pi as power supply with at least 2.5A, and it was connected with a ribbon cable to a display-port on the Raspberry Pi.

A Raspberry Pi Camera Module V2 (Model: RPI 8MP CAMERA BOARD) specially made to fit a Raspberry Pi was used to capture a image. It has a Sony IMX219 8-megapixel sensor and is capable to take images up to 3280$\times$2464. It was connected through a 15cm ribbon cable to a CSI-port on the Raspberry Pi. There are many third party libraries created for it and can therefore be used in Python with the Picamera Python library \cite{Picamera}.

As external soundcard, 
we employed the Plexgear USB-sound card USC-100.
As plug-in microphone, we used the Hama Mini Mikrofon Notebook Silver. 
It is a small microphone with a frequency range between 30Hz and 16 kHz, a sensitivity on -62 dB. and the impedance is 1kΩ.
The Roxcore Crossbeat Portable Bluetooth speaker (Model: Roxcore Crossbeat) was used as speaker.

All the hardware parts used in the project are shown in Figure~\ref{fig:hardware}, except the second Raspberry Pi which is attached to the backside of the display.
The company has its office on the second floor in a building, and the entrance is in the stairwell, therefore the devices was not exposed to rain, wind, sunlight etc. However there were some small echos in the stairwell, but that was not considered as loud enough to disturb the speech-to-text process. The lighting in the building was good enough for capturing a good picture, and there was no window near the entrance where the sunlight could cause a bad picture.

\subsection{Working Modes and GUI}

The system is divided in three functions for $i$) employees, $ii$) guests and  $iii$) deliveries. The employee function has two authentication steps, the face recognition and a random generated code that needs to be confirmed to protect against spoofing.
The guest function includes the speech-to-text service to state an employee's name that the guest wants to meet, and the employee is then notified.
The delivery function informs the specific persons in the office that are responsible for the deliveries by sending a notification.

The touch screen displays a GUI for the user and is built with the Tkinter library in Python. 
The main menu on the GUI shows the company logo and three buttons, one for each function.

\subsubsection{Employee Function}

The flowcharts of the employee function and GUI are shown in Figure~\ref{fig:flow_employee}.
If it is an employee, the system is going to ask the person visually through the GUI to stand in front of the camera to take a picture. When the picture has been captured, it is sent to the S3 bucket and deleted after the comparison. 
The image is encrypted on the client-side before it is uploaded to the cloud. The encryption algorithm used is the XOR cipher. XOR cipher is an easy used symmetric encryption algorithm that can be used for encrypting and decrypting images by changing the pictures byte arrays with a key, and this encryption algorithm is hard to crack by using the brute force method \cite{NATSHEH2016175}.
The image is decrypted before the comparison, and then compared against the collections using the Amazon Face Recognition API. To improve protection against spoofing attacks, a two factor authentication is implemented, by combination of face recognition and a code. When the comparison is done, it returns the similarity score. If the similarity score is above a predefined threshold, the system generates a random 4 digit code and sends it to the employee as a private message on Slack. This prevents unauthorized persons to enter the office. Even if the face recognition system recognizes the person on a picture, the person holding the picture would not be able to enter since the code is sent to the employee on the picture. The code needs to be entered on the keypad of the GUI. 
The system allows 3 unsuccessful attempts before it returns to the main menu, and then the process needs to be done again. Every new try, it generates a new random code to prevent a brute force attack. If the code is correct, the server unlocks the door, and sends a message to the GUI, which then displays a ``Welcome to the office'' message and the employee name.
If the similarity score is below the threshold, the person is denied access, and the GUI will go back to the main menu.

The lock system for the main door installed at the company works in a way such a voltage of 12V is constantly applied to the lock circuit to keep it closed. To activate (unlock) the door, a pulse signal of 0V has to be applied, and when the signal goes from 0V to 12V again, the door will unlock during 5 seconds.
In our prototype, this is implemented with the General Purpose Input/Output (GPIO) pins available at the Raspberry Pi.
For security, unlocking of the door is handled by the Raspberry Pi
placed  inside  the  office.

\subsubsection{Guest Function}

The flowcharts of the guest function and GUI are shown in Figure~\ref{fig:flow_guest}.
If the guest option is chosen, the person is asked visually through the GUI to pronounce the name of the employee that wishes to meet. The system records the input with the microphone and sends it to Google's Speech-to-Text service. 
Since it can be difficult to use speech-to-text on names, a string comparison algorithm is used to compare the input with all names in the database. If the most similar string has 
a similarity score over 80, the system sends a private message to the employee directly on Slack with the information that a guest is waiting outside the entrance door. The guest also gets information on the GUI that the employee have been notified. 
If the score is below 30, it then means that the similarity score is too low and the system will ask the user to try again. If the score is between these values, the guest needs to confirm that it is the correct person that they want to meet, with help of ``Yes'' and ``No'' buttons which will be displayed on the GUI. 
If the system returns the wrong employee name and the guest presses ``No'', the guest will be asked again who wishes to meet.

\subsubsection{Delivery Function}

The flowcharts of the delivery function and GUI are shown in Figure~\ref{fig:flow_delivery}.
If the delivery option is chosen, a notification is sent to the Slack channel that informs the employee in charge of deliveries that there is a delivery at the door. They will need to go to the door and open it manually since the delivery often requires a signature. 
The GUI will show that the employee has been notified.

\subsection{Data Protection and Maintenance}

Regarding General Data Protection Regulation (GDPR), the GUI employed allows to fulfill the requirement of the GDPR. According to GDPR, all kind of information that can directly be linked to a person, can be counted as personal data, even pictures, drawings and movies if they show a person. An audio file with a person's voice can also be considered as personal data \cite{Datainspektionen1,Datainspektionen2}. To avoid that the camera takes pictures of every person that stands in front of it, a button on the GUI for the employees is used, so it has to be activated on purpose. When a picture is taken, it is encrypted before it is sent for comparison, and then decrypted when it is going to be compared to images in the database. No pictures or audio are saved locally on the Raspberry Pi.

An strategy to keep images of the database updated is also employed. Only face images with a similarity score greater than a threshold (higher than 99.5\%) are used to update the template of the employee in the database. This also means that someone who is not an employee will have very difficult to obtain a sufficient score to have his/her picture saved, and the picture will be deleted immediately after the comparison.
The 10 most recent pictures of an employee are kept in the database. If additional images have to be stored, then the oldest images will be overwritten.

\begin{figure}[htb]
\centering
        \includegraphics[width=0.48\textwidth]{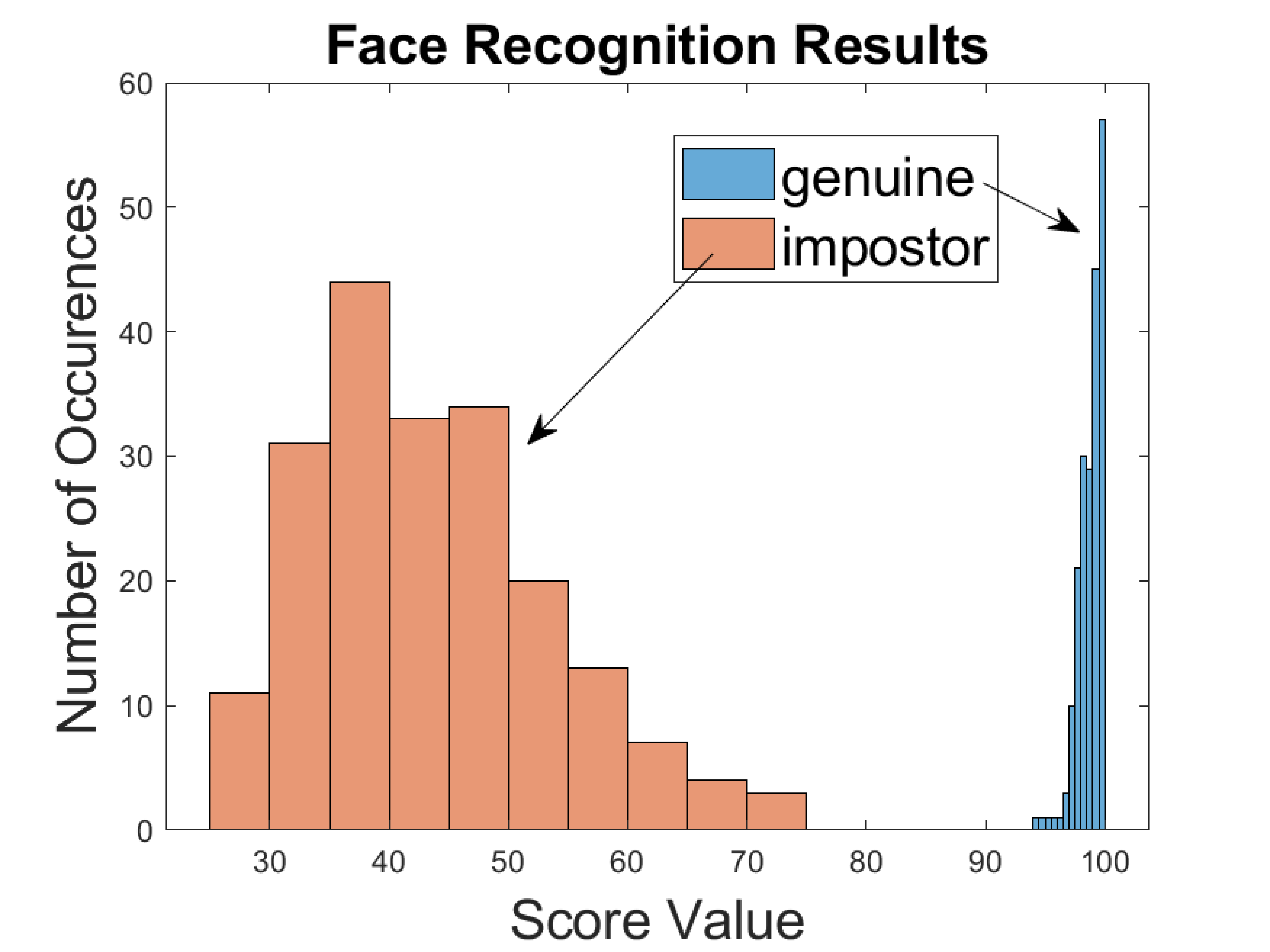}
\caption{Results of face recognition experiments.}
\label{fig:scores_histogram}
\end{figure}

\section{Experiments and Results}

The proposed prototype was tested to evaluate its
usability and accuracy. For this purpose,
the system was used by several staff members of the company.

\subsection{Face Recognition}

A total of 40 employees were used for the experiments.
To test the performance of the face solution employed, 20 employees were asked to carry out 10 separate genuine tests each, leading to 200 genuine trials.
Also, 5 unauthorized attempts were simulated against each of the 40 employees of the database, leading to 200 impostor trials.
The similarity score was saved for each recognition trial, with the
distributions of genuine and impostor scores plotted in Figure~\ref{fig:scores_histogram}. %
From our tests, Amazon Rekognition provided perfect separation between genuine and impostors. The smallest genuine score is 94.25\%,
while the highest impostor score is 73.1\%.
The system can be therefore considered reliable to allow employees to access the office, and to deny access to unauthorized people, at least at the scale evaluated here and in the imaging environment where the company operates.

\begin{figure}[t!]
\centering
        \includegraphics[width=0.48\textwidth]{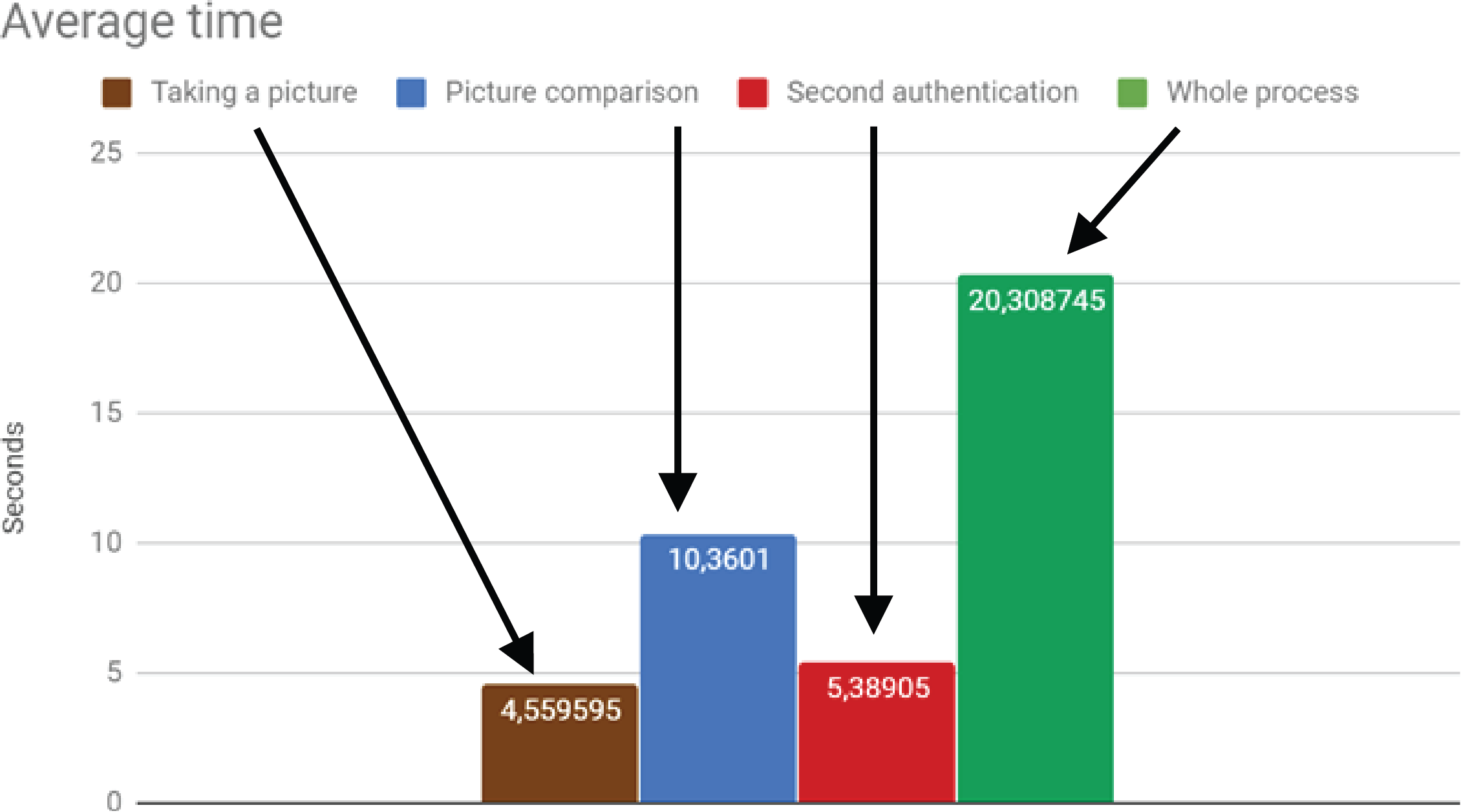}
\caption{Average time for the employee function in the system.}
\label{fig:average_time}
\end{figure}

During the genuine trials,
we also measured the time taken by the different necessary functions: capturing a picture, comparing it with the database, and introducing the random code. The whole operation covers from when the user presses the button in the GUI until the door is opened.
Average results are given in Figure~\ref{fig:average_time}.
Approximately 22\% of the time is consumed by taking the picture, 51\% by the authentication process carried out in the Amazon cloud, and 27\% by the introduction of the PIN sent to the employee. Half of the time is spent on comparing the image to the database, while the rest is divided between the other two processes in approximately equal parts.
Overall, the process takes 20.3 seconds in average.

\subsection{Speech-to-Text}

\begin{figure}[t!]
\centering
        \includegraphics[width=0.48\textwidth]{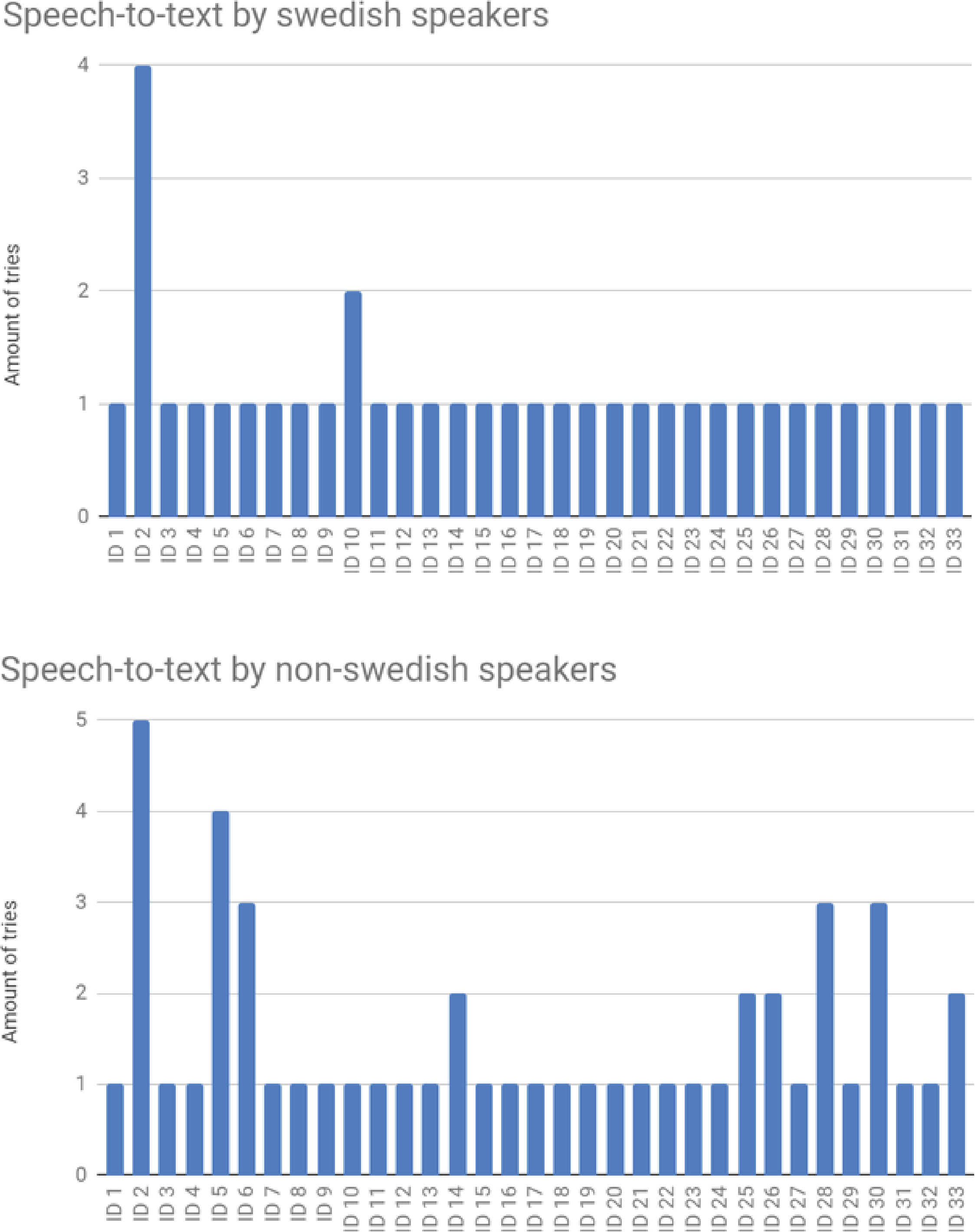}
\caption{Speech-to-text accuracy by swedish speakers (top) and non-Swedish speakers (bottom).}
\label{fig:stt}
\end{figure}

The language on the speech-to-text service was set to Swedish, and the tests intend to show how well the service works depending on if it is a native or a non-native Swedish speaker who is pronouncing. Each employee name has been spoken by the testers to see how many attempts was needed to match with the right employee. 
The results are shown in Figure~\ref{fig:stt}. As it can be shown, Swedish speakers managed to match the correct employee in the first try in the majority of cases (average amount of tries equal to 37/33= 1.12). On the other hand, non-Swedish speakers had to try more than once with approximately 27\% of the employees (9 out of 33 IDs), leading to an average amount of tries equal to 50/33=1.51.

\section{Discussion}

The goal of this work is to develop a system that can control the main entrance of an office by using face recognition and speech-to-text.
It makes use of two Raspberry Pi with a camera, a microphone, and a speaker.
Face recognition and speech-to-text conversion are done with the
cloud-based solutions provided by Amazon Web Services \cite{AmazonRek} and Google Speech-to-Text \cite{GoogleSpeechText}, respectively,
The system is complemented with a touch screen display, and
a graphical user interface (GUI) which presents the detected classes
to the user.

Disturbances are caused when a person inside the office needs to interrupt the work to open the door for an employee, a guest, or to receive delivery.
Face recognition minimizes the disturbances since no tag-keys are needed, and employees can enter the office even if they forget it, without anyone having to open the door manually for them.
The guest handling using speech-to-text also minimizes the interference, since it only notifies the employee in question which the guest declares in the system. The employee will receive a private message on Slack with the notification, so others at the office will not be disturbed. The delivery function also has its advantage, as it only sends a delivery notice on Slack to those who are responsible for receiving the deliveries.
GDPR is not an issue \cite{Datainspektionen1,Datainspektionen2} since the camera or the microphone are not recording continuously, but only when the corresponding function is activated in the GUI. In addition, none of the captured images or audio is saved, and the data is deleted immediately after it is being used for the intended purpose.
%
Another security measure is that data is encrypted before it is sent to the cloud.

The face recognition system works with very good accuracy in the environment of this work, as shown in the tests.
Spoof recognition is currently not supported in Amazon Rekognition, and the service does not include encryption/decryption of images either.
Using a two steps authentication process (face recognition combined with a confirmation code) makes the system more secure and provides protection against spoofing.
To avoid brute force attacks, a new random code is created every time, and it is only valid for three attempts.
A downside is that it is an extra step that takes about 5 extra seconds on average, which we will seek to overcome by including spoofing detection mechanisms \cite{6990726}.
The average time spent by an employee until the door unlocks is of 20.3 seconds, which  may be perceived as high, although it provides a secure and accurate method for access control.

The speech-to-text is set to Swedish and works well for Swedish-speaking persons.
For a multicultural company, it can be more difficult to use the speech-to-text service.
A problem can also occur if two or more employees has the same first and last name. The system will return the first occurrence in the database, which might not be the correct person the guest want to visit.
Our system is implemented in a relatively quiet environment, but
it can be hard to listen and translate in setups with
a lot of background noise or if several people speak at the same time.

Our prototype works well in an office environment, but if it is connected to a slower internet connection, the face recognition and speech-to-text will take longer. 
Also, the system is cloud based, which means that it always requires internet connection to operate.
There are also advantages in using a GUI. It allows to easily show information for the person using the system, and also to instruct the person on how to enter the office. Based on the feedback obtained, it is more comfortable to visualize the information for both the employee and the visitor.

\section*{Acknowledgment}

Authors K. H.-D. and F. A.-F. thank the Swedish Research Council for funding their research.
Authors also acknowledge the CAISR program of the Swedish Knowledge Foundation.


\bibliographystyle{IEEEtran}

\bibliography{bibliography}

\end{document}